\documentclass[10pt, a4paper]{article}
\usepackage{lrec2022} 
\usepackage{multibib}
\newcites{languageresource}{Language Resources}
\usepackage{graphicx}
\usepackage{tabularx}
\usepackage{soul}
\usepackage{booktabs,arydshln}

\usepackage{titlesec}
\titleformat{\section}{\normalfont\large\bfseries\center}{\thesection.}{1em}{}
\titleformat{\subsection}{\normalfont\SmallTitleFont\bfseries\raggedright}{\thesubsection.}{1em}{}
\titleformat{\subsubsection}{\normalfont\normalsize\bfseries\raggedright}{\thesubsubsection.}{1em}{}
\renewcommand\thesection{\arabic{section}}
\renewcommand\thesubsection{\thesection.\arabic{subsection}}
\renewcommand\thesubsubsection{\thesubsection.\arabic{subsubsection}}

\usepackage{epstopdf}
\usepackage[utf8]{inputenc}

\usepackage{hyperref}
\usepackage{xstring}
\usepackage{fancyhdr,graphicx,amsmath,amssymb}

\usepackage[ruled,vlined]{algorithm2e}
\usepackage{xcolor}
\usepackage{graphicx}
\usepackage{adjustbox}
\usepackage{multicol}
\usepackage{multirow}
\usepackage{xurl}
\usepackage{booktabs}

\newcommand{\ie}[0]{\textit{i.e.},\ }   
\newcommand{\eg}[0]{\textit{e.g.},\ }   

\title{TWEET-FID: An Annotated Dataset for Multiple Foodborne Illness Detection Tasks}

\let\SUP\textsuperscript

\name{Ruofan Hu\SUP{1}, Dongyu Zhang\SUP{1}, Dandan Tao\SUP{2}, Thomas Hartvigsen\SUP{3}, Hao Feng\SUP{4}, Elke Rundensteiner\SUP{1}} 

\address{ \SUP{1}Data Science Program, Worcester Polytechnic Institute, Worcester, MA, USA \\
\SUP{2}Vanke School of Public Health, Tsinghua University, Beijing, China \\
\SUP{3}CSAIL, Massachusetts Insititute of Technology, Cambridge, MA, USA\\
\SUP{4}Dept. of Food Science \& Human Nutrition, Univ. of Illinois, Urbana-Champaign, Urbana, IL, USA \\
         {\{rhu, dzhang5, rundenst\}@wpi.edu}, tomh@mit.edu, {\{dtao2, haofeng\}@illinois.edu}\\}

\begin{document}
\abstract{
Foodborne illness is a serious but preventable public health problem -- with delays in detecting the associated outbreaks resulting in productivity loss, expensive recalls, public safety hazards, and even loss of life. While social media is a promising source for identifying unreported foodborne illnesses, there is a dearth of labeled datasets  for developing effective outbreak detection models. To accelerate the development of machine learning-based models for foodborne outbreak detection, we thus present TWEET-FID (TWEET-Foodborne Illness Detection), the first publicly available annotated dataset for multiple foodborne illness incident detection tasks. TWEET-FID collected from Twitter is annotated with three facets: tweet class, entity type, and slot type, with labels produced by experts as well as by crowdsource workers. We introduce several domain tasks leveraging these three facets: text relevance classification (TRC), entity mention detection (EMD), and slot filling (SF). We describe the end-to-end methodology for dataset design, creation, and labeling for supporting model development for these tasks. A comprehensive set of results for these tasks leveraging state-of-the-art single- and multi-task deep learning methods on the  TWEET-FID dataset are provided. This dataset opens opportunities for future research in foodborne outbreak detection. 
\\ \newline  }

\maketitleabstract

\section{Introduction}
Foodborne illnesses continue to threaten public health. Approximately 1 in 6 Americans (or 48 million people) are sickened by foodborne illness each year \cite{scallan2011foodborne}. Foodborne illnesses lead to loss of productivity, medical expenses, and even death. The annual economic cost caused by foodborne illnesses in the United States is estimated to be between \$14 to  \$60 billion \cite{hoffmann2020acute,scharff2018economic}.

Detecting foodborne illnesses early promises to reduce the risk and curtail the outbreak. In the past decade, it has been recognized that user-generated public posts on social media or review platforms can play an important role in the surveillance of unreported cases of foodborne illness; Surveillance applications are being piloted by local health agencies, including mining data from Twitter in New York City \cite{harrison2014using}, 
Las Vegas \cite{sadilek2016deploying}, 
and consumer review sites such as Yelp in San Francisco \cite{schomberg2016supplementing} and New York City \cite{effland2018discovering}.
However, these applications tend to focus only  on a
'post-level' inspection, where a followed-up labor-intensive manual phase is deployed to extract detailed information from potential posts related to foodborne illness after they were selected through machine learning models.
This procedure
is not only slow and costly,
but also risks missing potentially-valuable pieces of information.
\begin{figure*}[th]
\centering
\includegraphics[width=\linewidth]{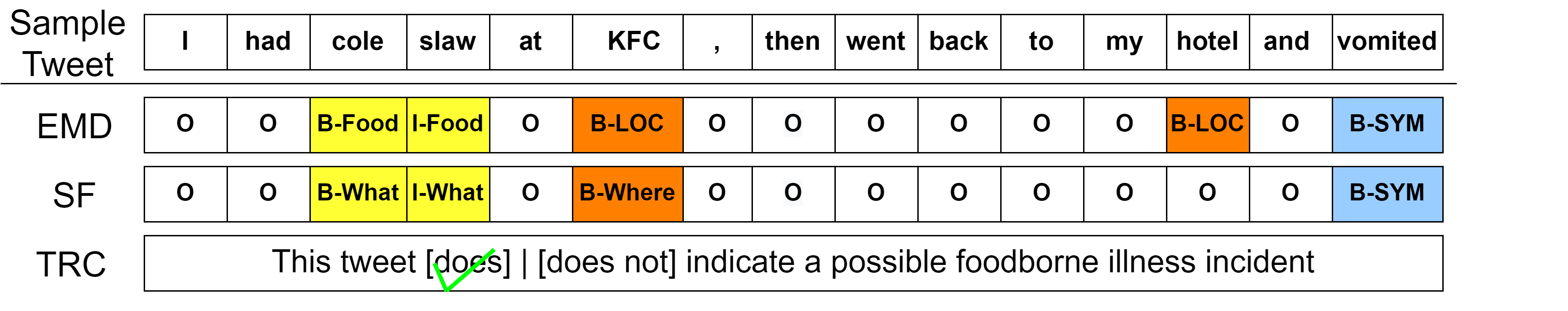}
\caption{An example tweet (top) with labels for several tasks in our TWEET-FID dataset (bottom 3 rows). EMD: Entity Mention Detection, SF: Slot Filling, TRC: Text Relevance Classification.
\underline{Entity Type Abbreviations}: LOC: Location, SYM: symptom.
For TRC, the green checkmark indicates  the correct task label.
}
\label{fig:sample_tweet}
\end{figure*}
Instead, a surveillance system should automate extracting as much foodborne illness incident-specific information as possible. A successful method would not only determine if a given tweet indicates a possible foodborne illness incident, but also automatically extract critical entities from the tweet so that they can be aggregated into trends of sufficient size to be acted upon. We accelerate this tweet inspection process, dividing it into three tasks that can be automated: 
(1) identify if the tweet indicates a foodborne illness incident; (2) find and extract mentions of food, symptoms, locations, and foodborne illness-related keywords in the tweet; and (3) recognize mentions of slots (\ie the attributes of one incident like \textit{What},\textit{Where})  and their values related to any foodborne illness incident described in the tweet. These tasks are text relevance classification (TRC), entity mention detection (EMD), and slot filling (SF), respectively.  

As an example, shown in Figure \ref{fig:sample_tweet}, 
consider the 
tweet: \textit{"I had a \textbf{cole slaw} at \textbf{KFC}, then went back to my \textbf{hotel} and \textbf{vomited}"}. 
This tweet indicates a possible foodborne illness incident, where \textit{\textbf{cole slaw}} could be the contaminant, \textit{\textbf{KFC}} the place where the person had the contaminated food, and \textit{\textbf{vomit}} the symptom of food poisoning. While the \textit{\textbf{hotel}} is also a location, it plays no role in the poisoning. Information about the mentions of slots and their values related to the incident can help regulators, government officials, and companies trace the causes and effects of foodborne illnesses. 


Our work introduces a publicly-available and carefully-curated dataset, TWEET-FID (TWEET-Foodborne Illness Detection), designed by food science experts to support these three tasks by covering multilevel information about foodborne illness incidents. 
As depicted in Figure \ref{fig:sample_tweet}, 
our dataset provides tweets with multiple task labels that can be leveraged for training models for the TRC, EMD, and SF tasks described above. 
To develop TWEET-FID, food safety experts in our team retrieved English posts from Twitter based on an extended keyword list established in \cite{effland2018discovering}. They selected the keywords with which we retrieve as many potential relevant tweets as possible (\ie have a high recall). Unfortunately, improving recall tends to hurt precision, so the dataset may inevitably also include many irrelevant tweets. 


Compared to expert-given labels, crowd-sourced labeling can be cost-effective and time-efficient. 
However, a main concern of crowd-sourcing is annotation quality. To clear this concern and develop a reliable and general method of large-scale dataset construction, we first deploy crowd-sourcing to generate multiple annotations for each tweet covering the three facets above. We then establish a quality control mechanism to 
clean these crowd-sourced annotation sets based on high inter-annotator agreement. Even further, we then ask a group of in-house food scientists to provide some annotations as well. By evaluating the filtered crowd-sourced annotations with the unified expert label as the ground-truth label, we succeed in confirming the effectiveness of our quality control mechanism. Our annotation approach can thus be leveraged for similar future work on dataset construction.  

To label our tweets, annotators are asked to complete three subtasks: rate the tweet on a Likert scale, tag all relevant entity types in the tweet, and subsequently decide whether the selected entity mention is the value to fill into the designated slots. 
We enforce that at least one slot must be filled whenever the tweet indicates a possible foodborne illness incident. 
Then, we create an aggregated class label per tweet using majority voting (MV) and apply two methods, BSC-seq \cite{simpson-gurevych-2019-bayesian} and MV, to aggregate the entity mention labels and slot type labels. The evaluation results indicate that majority voting generates acceptable aggregated labels for all three task labels, while more advanced algorithms are needed for token-level annotation aggregation.


Given our labeled TWEET-FID dataset, we then conduct a comprehensive experimental study of deep learning models performing the three proposed tasks: TRC, EMD, and SF. We utilize the aggregated labels in Tweet-FID to train state-of-the-art single- and multi-task deep learning models such as RoBERTa \cite{liu2019roberta}, BiLSTM \cite{graves2013speech}, and MGADE \cite{wunnava-etal-2020-dual}. We then evaluate the trained models with the test set with the ground truth label. 

Experimentally, we observe that multi-task methods match or exceed single-task methods, likely by learning from interrelationships between the tasks. 
Our experimental findings illustrate that effective models for these critical tasks can indeed be found. Furthermore, our dataset also provides the opportunity for future research studies to use our work as a baseline. 

Our contributions are as follows:
\begin{itemize}

    \item {We developed TWEET-FID, the first publicly-available social media dataset with multiple task labels, representing a valuable resource for model development in foodborne disease surveillance.}
    
    \item {TWEET-FID covers three facets: (i) presence of foodborne illness, (ii) entity types in each tweet, and (iii) slot types: the semantic role of the entity mention in the tweets indicating foodborne illness incident. We create aggregated labels for each fact by crowd-sourcing annotations and finding high inter-annotator agreement.} 
    
    \item {In addition, we created ground truth labels based on the annotations performed by trained in-house annotators. This allowed us to  confirm that the aggregated labels are of high quality. Our work provides a successful methodology for constructing labeled  datasets for multi-grained information extraction.}

    \item {We design an experimental study that realizes the above three domain tasks by implementing them using  single-task and multi-task supervised deep learning-based models on the TWEET-FID dataset. Our findings provide a much-needed reference point for research in model development in foodborne disease surveillance.}
\end{itemize}

\section{Related Work}
Social media data has been identified as a great source of information for public health due to its timeliness and scalability. 
Many researchers have built effective surveillance systems in the food safety domain, even with vanilla machine learning methods. Most previous work only collect one class label per tweet (\ie whether it is relevant to foodborne illness events). They identified sick Yelp reviews or tweets with simple machine learning models (like logistic regression, SVM, and Random Forest) for a particular area 
\cite{harrison2014using,harris2014health,harris2017research,schomberg2016supplementing,sadilek2016deploying,effland2018discovering}. While these prior works classify the relevant posts successfully, more detailed information is retrieved manually during the inspection process, as explained in Section 1. \newcite{haotian2021integrated} constructed a Chinese corpus of food safety incident entities using Bi-LSTM-CRF from food safety incident-related news. In our work, we collected not only the sentence label but also the entities mentioned in English tweets and the semantic role they play in foodborne illnesses.

Crowdsourcing is a good option for collecting labels for new datasets at scale \cite{paun2021aggregating}. Some works have collected tweets on public health topics and used crowdsourcing to collect annotations. For instance, \newcite{ghenai2017catching} construct a twitter dataset to identify Zika-related rumors around the world. Crowdsourcing workers are used to annotate tweet with at least three labels, in which majority voting determines the aggregated labels. However, they did not evaluate the final aggregated labels.  \newcite{zong2020extracting} collect tweets for Covid-19 event extraction and have at least seven crowdsourcing workers perform annotations on event type and slot values for each tweet. Then, the gold labels for all slots and events are generated by majority votes. They design a quality control mechanism by comparing workers' annotations with the majority votes and filtering out the workers' F1 scores below 0.65. The evaluation results of the aggregated labels of 100 tweets show that the aggregated labels are suitable. However, the small set of 100 tweets cannot draw a firm conclusion that either the quality control mechanism or the aggregation methods are really effective enough to construct a good quality dataset for such a critical public health-related study. In this work, we collect crowdsourcing labels with a quality control mechanism, create aggregated labels, and provide ground truth labels for the whole dataset to evaluate the effectiveness of the quality control mechanism and the chosen label aggregation methods.


\section{Dataset Creation}
It is challenging to collect multi-grained annotations for a specific domain. First, the tasks should be illustrated clearly to collect high-quality labels. Furthermore, it is even harder to merge the labels from multiple annotators into an integrated label. In this section, we describe the data collection process, show the ground truth label generation from the expert labels, provide the annotation strategies, and present the data aggregation procedure.

\subsection{Data collection and preprocessing} \label{sec:data_collection}
Using the Twitter API \footnote{\url{https://developer.twitter.com/en/docs/twitter-api}}, we use keywords to search for tweets from as early as January 2019. As recommended by food safety experts and the literature \cite{effland2018discovering} our keywords include common terms and hashtags indicative of foodborne illness: \textbf{''\#foodpoisoning''}, \textbf{''\#stomachache''}, \textbf{''food poison''}, \textbf{''food poisoning''}, \textbf{''stomach''}, \textbf{''vomit''}, \textbf{''puke''}, \textbf{''diarrhea''}, and \textbf{''the runs''}. We ended up collecting a total of 6,192,537 tweets and filtered out retweets and those with fewer than four tokens to clean the dataset. To develop an effective methodology for future label collection via crowdsourcing at large-scale, we further select 4122  tweets (including 1000 tweets without mentions of these keywords to diversify the dataset) to be labeled by crowd workers and trained experts. This dataset is relative balanced in regard to foodborne illness incident or not. 
We then normalize the tweets by converting user mentions and URL links to \verb+@USER+, \verb+HTTPURL+, respectively. Since emojis may carry important information, we keep emojis and translate the icons into text strings by using the \verb+emoji+ module\footnote{\url{https://pypi.org/project/emoji/}}.

\subsection{Crowdsourced annotation}
\begin{figure*}[t]
\centering
\includegraphics[width=0.75\linewidth]{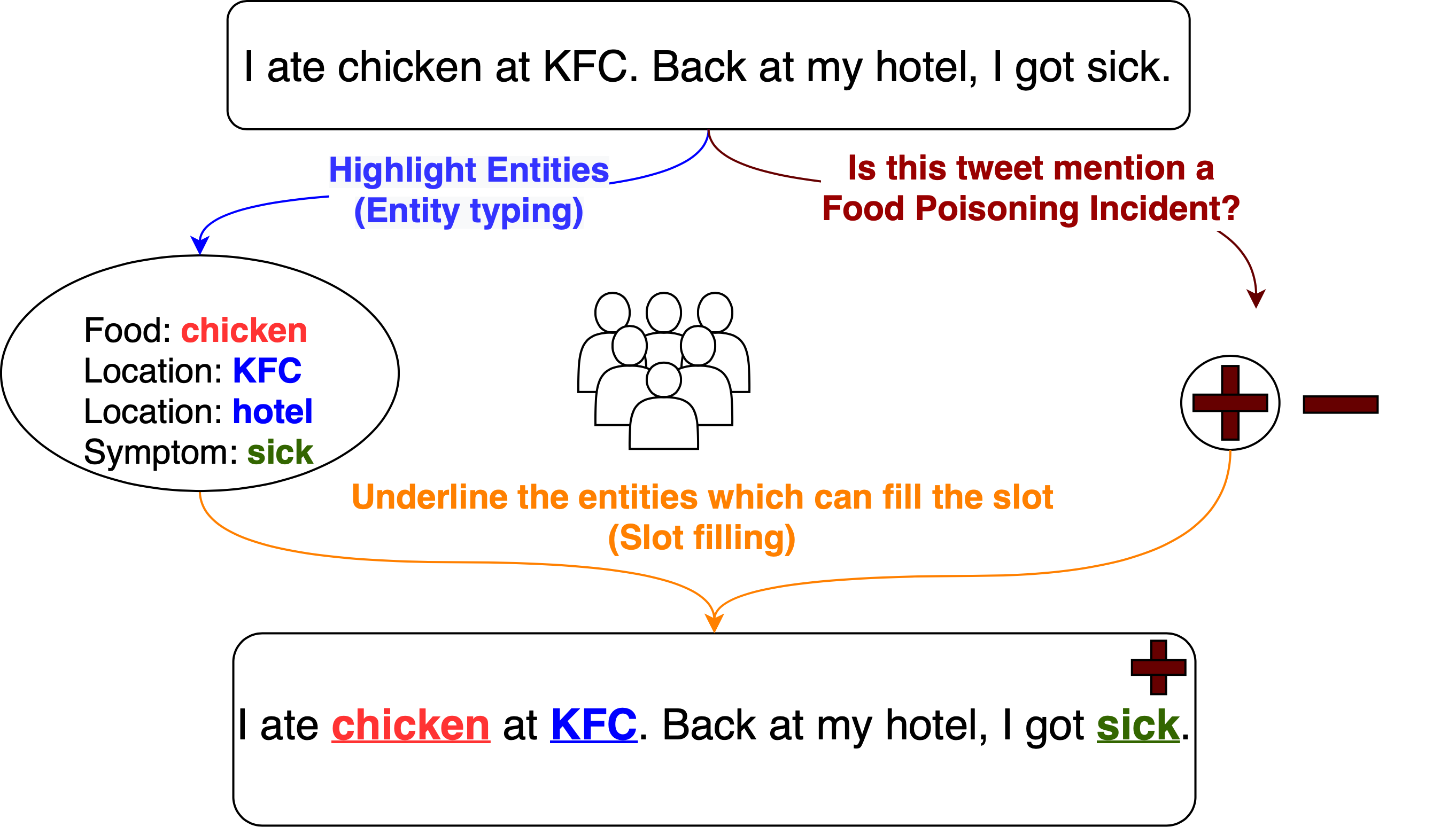}
\caption{The workflow of annotating.}
\label{fig:flow}
  \vspace{-10pt}
\end{figure*}

We use the Amazon Mechanical Turk (MTurk) crowdsourcing platform to elicit multiple annotations per tweet with a custom interface. An example of the annotating workflow is shown in Figure \ref{fig:flow}.

Participants were asked to complete three tasks: 1) rate\footnote{We kept this part of the design during our data collection phase. It is useful for either build a regression model or classification model after converting the rating to a binary label.} the tweet by whether they think it indicates a foodborne illness incident on a Likert scale of 0-5, 2) Highlight all words/phrases that are mentions of food, location, symptoms, and foodborne illness keywords, and 3) Indicate for each highlighted words to fill the slot defined in Table \ref{tab:slot_def}. \begin{table}[!htb]
\begin{center}
\begin{tabularx}{\columnwidth}{|c|X|}

      \hline
      \textbf{Slot} & \textbf{Definition}\\
      \hline
      What & The food have caused the possible foodborne illness incident to happen.\\
      \hline
      Symptom & The symptoms that the person had.\\
      \hline
     Where & The location the person buy the food.\\
      \hline
     Key & Other keywords related to foodborne illness: e.g., food poisoning.\\
     \hline
\end{tabularx}
\caption{Slot definition used for annotation.}
\label{tab:slot_def}
  \vspace{-10pt}
 \end{center}
\end{table}

\subsubsection{Pilot study} \label{sec:pilot_study}

{\bf In-house study for the design of annotation instructions and the interface.} To ensure the usability and clarity of our labeling interface and instructions, we begin with an in-house study.
In our preliminary user interface, we provide verbal instructions that explain the project objective, display the screenshot of one completed annotation to help the annotators understand the task, and show the definition of each slot in a foodborne illness incident. We asked 12 graduate students to participate in this study. Participants asked to complete the three annotation tasks shown in Figure \ref{fig:flow} for several tweets. Then we asked them for constructive feedback and suggestions on the interface design. It turned out that they were not familiar with the expected annotations nor the process of providing annotations for the first time. To overcome this, we embedded a two-minute video into the interface with a detailed explanation of each task as well as showcasing each of the manipulations available to the annotators which turned out a 0.98 sentence-level annotation agreement and the inter-annotator agreement Krippendorff's ${\alpha}_{u}$ \footnote{https://github.com/LightTag/simpledorff } \cite{krippendorff2011computing} of the entity type annotation and slot type annotation are 0.854 and 0.744 respectively.

{\bf Investigating annotation quality.} Next, we conduct a pilot study to investigate workers' performance on MTurk. Although MTurk provides a mechanism to only allow workers with a high approval rate to take a task, this screening does not reliably filter out workers who may provide low-quality annotations for our complicated task. Further, using more annotators per instance can help identify low-quality work \cite{zheng2017truth}.
Thus, we set out to develop an effective quality control mechanism while also calibrating an equitable payment strategy. For this, we published an initial batch of 200 tweets. We set the reward as ${\$}$0.1 per. 
HIT\footnote{HIT refers to a single, self-contained, virtual task that a worker can work on, submit an answer, and collect a reward for completing. 
In our case, the three questions 
 shown in the same interface is a HIT.
}, 
based on an estimation of a reasonable working time of roughly 50 seconds per tweet \footnote{This corresponds to the US federal minimum wage ({$\$$}7.25/hour).}. Each tweet was assigned to five unique annotators, resulting in a total of 1000 annotations collected from 110 annotators. Intuitively, the task of tagging entities requires more effort from annotators than rating tweet relevance. Furthermore, identifying slot values relies on effective entity tagging. 
Thus, we focused on analyzing the entity annotation task to filter out low-quality work that may negatively affect the quality of label aggregation.
We measure the similarity over all pairs of crowd annotations with an average macro-F1 score, and the complete algorithm is shown in Algorithm \ref{tab:pseudocode}.
We keep only tweets with an average F1 score above a threshold of 0.7. The inter-annotator agreement Krippendorff's ${\alpha}_{u}$ was raised from 0.35 to 0.61 after dropping these low quality annotations. \begin{algorithm}[thb]
\SetAlgoLined
\KwResult{A list of the rating for each annotation of one tweet}
 A=$\{a_{1},a_{2},\ldots,a_{m}\}$, $\overline{R}=[\ ]$ \;
 \For{$a_{i} \in A $}  {
  $R_{i}=[\ ]$ \;
  \For {$a_{j} \in A, j{\neq}i  $}{
  $ r_{i,j} = similarity (a_{i},a_{j})$ \;
  $ R_{i}.append( r_{i,j} )$ \
  }
  $ \overline{R} .append(\ R_{i}/{len(R_{i})} )$ \
 }
$\overline{R} \gets {\overline{R}/max(\overline{R}) }$
\caption{Rating for entity type annotations of one tweet}
\label{tab:pseudocode}
\end{algorithm}

{\bf Working time and compensation.} The average working time in the in-house study was 15s. Thus we deemed 50s on average as a reasonable time for the workers and set the corresponding compensation to ${\$}$0.1 per HIT for the first pilot. The results of this pilot study showed an average working time of 47s for good-quality jobs - confirming our compensation strategy. 

{\bf Controlling annotation quality.} In the first pilot study, two of the 110 annotators took more than 80 HITs on MTurk, but over 50${\%}$ of their work was identified as low quality. Thus, we need to address this risk that a few workers who perform low-quality work may do a great deal of damage to the quality of the final generated labels. To assure the quality of the collected annotation set, our strategy is to split our dataset into small batches and control the  HITs available for a worker in each batch - allowing us to confirm their agreement with other workers iteratively. The improvement of the inter-annotator agreement confirms the effectiveness of our quality control mechanism for identifying and preventing low-quality work. 

\subsubsection{Main study} 
Based on our findings from the pilot study above, we then conduct the main study as follows. For each tweet, we collect annotations for five workers. We rejected low-quality work which is determined with Algorithm \ref{tab:pseudocode}, re-assigning the rejected annotations to other workers. Since the rejection may affect a worker's reputation, we temporarily rejected the low-quality work and approved them after obtaining at least 5 high-quality annotations. We divided the whole dataset into eight batches, and workers were only allowed to take at most 10 tasks in each batch. This way, we can both alleviate the temporary negative impact of our rejection on the annotators and concurrently reduce the potential bias in the collected annotations. If many jobs submitted by a worker were identified as being low-quality, we assign a qualification to the worker to flag this internally within our system and they are no longer qualified to perform more of our tasks. We publish a new batch only after each tweet in the previous batch has a sufficient number of good annotations. 

\subsection{Crowdsourced Label aggregation} \label{sec:aggregated}
We next create aggregated labels for the entity types and overall tweet in parallel since the candidate slot fillers must be entity mentions and the defined slots only exist in the tweets indicating foodborne illness. Subsequently, given the entity label and the sentence label, we can create the slot type label. Since entity typing task and slot filling are both sequence tagging tasks, we employ two methods, majority voting(MV) and Bayesian sequence combination (BSC) referred to as \textit{\textbf{BSC-seq}} in \cite{simpson-gurevych-2019-bayesian}, to aggregate the sequence labels and evaluate the aggregation results in Section \ref{sec:dataset description} to determine which method is more reliable to be used in future similar study on sequence annotation aggregation. 

{\bf Tweet class label.} The class label (relevant to foodborne illness or not) of a tweet is decided based on majority voting. Since we collected the sentence score on a scale of 0-5, we need to convert the rating score to binary label \ie we transformed the rating score to the class label 1 if it is higher than 2, and class 0 otherwise. Then we applied majority voting on the binary labels (ties are assigned to class 1). 

{\bf Entity type label. } To represent the beginning, inside, and outside of each entity, we convert the annotations into labels with the \textit{BIO} schema  (BIO stands for Beginning, Inside and Outside of an entity). The majority voting results may include some labels against the rules, \eg the label of an entity may begin with 'I-'. For these cases, we assign a label of "B-entity$\_$type" to the token before. For the BSC method, we do no postprocessing on the aggregated labels since this method is specifically for sequence annotation aggregation.

{\bf Slot type label.} If the tweet is irrelevant to a foodborne illness incident, we ignore the slot type annotations of all entities included in the tweet. Otherwise, we first drop the annotations with inconsistent entity boundary or entity labels with the aggregated entity label. Then, we employed majority voting and BSC to determine the slot type on the left annotations separately.

\subsection{Generate ground truth from expert labels}
We collect two expert labels for each tweet from 12 trained graduate students who major in either data science or food science and then unify them as the ground truth label. For the tweet class label, since we collected the Likert label of the tweet, we need to convert the rating score to a binary label, which we achieve by transforming the rating score to the class label 1 if it is higher than 2, and class 0 otherwise. We gather a third person to make a final decision if the labels of the two sentences cannot reach an agreement. For the entity type, we take the union of the two expert entity type labels as the final ground truth. Subsequently, we take the union of the two expert labels of slot type and solve the conflict with the help of a third person. 
There are only 86 tweets (${\sim}$2$\%$) that have conflicting expert annotations on the sentence label. 
As for the slot filling label, only 307 (${\sim}$7$\%$) tweets, including the 86 tweets with conflict sentence-level annotations, have conflict expert annotations.
This confirms that the experts in the vast majority of the cases have a strong  agreement on the labels for the most of the instances. Furthermore, we can thus confidently conclude that there is a 'ground truth' for the dataset.

\section{Dataset Characteristics}\label{sec:dataset description}

\subsection{Dataset splits and statistics}
Our dataset includes 1362 (33$\%$) tweets relevant to the spread of foodborne illness and 2760 (67$\%$) irrelevant tweets. 758 tweets among the irrelevant ones do not contain any entities. We do a train-validation-test split for Tweet-FID based on the unified expert label. The training set consists of 1088 relevant tweets and 2210 irrelevant tweets. The validation set includes 137 relevant tweets and 275 irrelevant tweets. The test set consists of 137 relevant tweets and 275 irrelevant tweets. Thus,  the split ratio is close to 8:1:1. This split is stratified by the tweet-level relevance class. Detailed statistics of the entity type labels and slot type labels are shown in the Appendix.

\subsection{Quality Evaluation} 

\subsubsection{Evaluation metrics} \label{sec:annotation eva}
The three annotation tasks are classical NLP: either sentence classification or sequence labeling. We use accuracy to measure the tweet class labels. The entity type and slot type correspond to sequence labeling problem, thus we evaluate these labels with \textbf{Strict ${F_1}$} and \textbf{Type ${F_1}$}\footnote{\url{https://pypi.org/project/nervaluate/}}, both of which are evaluation metrics introduced in semEval-2013-task9. A named entity usually has a clear boundary, however, people may segment the entity mentions at different positions. For example, \textit{"my stomach hurts really bad."} describes the symptom and some people may only take \textit{"stomach hurts"} but others may select \textit{"stomach hurts really bad"} as the symptom. All of them have identified the text span successfully, but the boundaries do not exactly match. Under such a scenario, the \textbf{Type ${F_1}$} is more appropriate as a metric.

\subsubsection{Quality of the crowdsourced annotations}\label{sec:quality_crowd}
We evaluate the five crowdsourced annotations on the three tasks for each instance by referring to the unified expert label as the ground truth. The evaluation results are shown in Table \ref{tab:crowd_5_eval}. We find that the annotation quality of the train/validation/test is similar, and the train/validation/test sets have similar data distribution. There are 20,610 annotations collected from 3730 workers, so each worker contributes 5.5 annotations on average, and the median value is 5.
\begin{table}[!htb]

\begin{adjustbox}{width=\linewidth}
{
\huge
    \centering
    \begin{adjustbox}{width=\linewidth}
        \begin{tabular}{l c c c c c c} 
        \toprule
        \multirow{2}{4cm}{} & \multicolumn{2}{c}{Sentence Label} & \multicolumn{2}{c}{Entity Type} & \multicolumn{2}{c}{Slot Type} \\
		\cmidrule(l{3pt}r{3pt}){2-3} \cmidrule(l{3pt}r{3pt}){4-5} \cmidrule(l{3pt}r{3pt}){6-7} 
		& Accuracy  & ${F_1}$ & Strict ${F_1}$ & Type ${F_1}$ & Strict ${F_1}$ & Type ${F_1}$ \\
		\midrule[\heavyrulewidth]

Train   & 0.745     & 0.687         & 0.637      & 0.715 & 0.477 & 0.534   \\
Validation   & 0.766     & 0.711         & 0.634      & 0.714    & 0.478 & 0.540   \\
Test   & 0.724     & 0.679         & 0.578      & 0.666    & 0.435 & 0.505   \\
\midrule
\textbf{Full set}   & 0.744     & 0.689         & 0.631      & 0.710    & 0.472 & 0.531   \\
		\bottomrule
		\end{tabular}
		\end{adjustbox}

}  
\end{adjustbox}
  \caption{Evaluation results of all annotaions for the three labels of sentence label, entity type, and slot type. Metrics: Accuarcy and ${F_1}$ for sentence label, Strict ${F_1}$ and Entity Type ${F_1}$ for entity type and slot type.\\ }
  \label{tab:crowd_5_eval}
  \vspace{-10pt}
\end{table}

\subsubsection{Quality of the aggregated label} We evaluate the aggregated label for the training and validation set by comparing it with the ground truth since the label quality of the training and validation set may affect the downstream learning tasks. The evaluation results in Table \ref{tab:aggregated_eval} confirm the effectiveness of the quality control mechanism and aggregation methods for entity type labels. One reason for the relatively worse performance of BSC is the \textbf{sparsity} of the worker annotation. As mentioned in Section \ref{sec:quality_crowd}, each worker only gives 5.5 annotations on average and the median value is 5, which means it is difficult for the annotator model to learn the noise and bias of each annotator due to the limited information. The aggregation methods for slot type label aggregation are non-ideal, and more advanced methods are needed for the challenging slot type annotation aggregation task.
\begin{table}[!htb]

\begin{adjustbox}{width=\linewidth,center}
{
\huge
    \centering
    \begin{adjustbox}{width=\linewidth,center}
        \begin{tabular}{l c c c c c c} 
        \toprule
        \multirow{2}{4cm}{Method} & \multicolumn{2}{c}{Sentence Label} & \multicolumn{2}{c}{Entity Type} & \multicolumn{2}{c}{Slot Type} \\
		\cmidrule(l{3pt}r{3pt}){2-3} \cmidrule(l{3pt}r{3pt}){4-5} \cmidrule(l{3pt}r{3pt}){6-7} 
		& Accuracy  & ${F_1}$ & Strict ${F_1}$ & Type ${F_1}$ & Strict ${F_1}$ & Type ${F_1}$ \\
		\midrule[\heavyrulewidth]

MV   & 0.824     & 0.791         & \textbf{0.784}      & \textbf{0.828}    & \textbf{0.613} & \textbf{0.657}   \\
BSC   & NA     & NA         & 0.744      & 0.823    & 0.484 & 0.531   \\

		\bottomrule
		\end{tabular}
		\end{adjustbox}

}  
\end{adjustbox}
  \caption{Evaluation of the aggregation for the three labels of sentence label, entity type, and slot type on the training and validation sets.}
  \label{tab:aggregated_eval}
  \vspace{-10pt}
\end{table}

\section{Experiments}
\subsection{Evaluation tasks}
\label{sec:tasks}

\begin{table*}[htb]

\begin{adjustbox}{width=0.95\linewidth,center}
{
\huge
    \centering
    \begin{adjustbox}{width=\linewidth,center}
        \begin{tabular}{l l c c c c c c} 
        \toprule
        \multirow{2}{4cm}{Label Source} &\multirow{2}{7cm}{Learning Methods} & \multicolumn{2}{c}{TRC} & \multicolumn{2}{c}{EMD} & \multicolumn{2}{c}{SF} \\
		\cmidrule(l{3pt}r{3pt}){3-4} \cmidrule(l{3pt}r{3pt}){5-6} \cmidrule(l{3pt}r{3pt}){7-8} 
		&&Accuracy & ${F_1}$ & Strict ${F_1}$ & Type ${F_1}$ & Strict ${F_1}$ & Type ${F_1}$ \\
		\midrule[\heavyrulewidth]

\multirow{12}{*}{MV}

& RoBERTa - TRC & .730 ± .013 & .707 ± .011 & N/A & N/A & N/A & N/A\\
& RoBERTa* - EMD & N/A & N/A & .729 ± .003 & .804 ± .003 & N/A & N/A\\
& RoBERTa* - SF & N/A & N/A & N/A & N/A & .565 ± .003 & .621 ± .009\\
& RoBERTa* - TRC+EMD & .778 ± .018 & .743 ± .014 & .745 ± .007 & \textbf{.811 ± .005} & N/A & N/A\\
& RoBERTa* - TRC+SF & \textbf{.795 ± .021} & \textbf{.755 ± .018} & N/A & N/A & \textbf{.599 ± .012} & \textbf{.664 ± .015}\\
& RoBERTa* - EMD+SF & N/A & N/A & \textbf{.747 ± .004} & .809 ± .008 & .597 ± .011 & .651 ± .014\\

\cdashline{2-8}[2pt/3pt]
& BiLSTM - TRC & .720 ± .007 & .677 ± .011 & N/A & N/A & N/A & N/A\\
& BiLSTM* - EMD & N/A & N/A & .705 ± .005 & .766 ± .006 & N/A & N/A\\
& BiLSTM* - SF & N/A & N/A & N/A & N/A & .538 ± .008 & .584 ± .015\\
& BiLSTM* - TRC+EMD & .744 ± .024 & .710 ± .017 & .716 ± .003 & .770 ± .006 & N/A & N/A\\
& BiLSTM* - TRC+SF & .751 ± .034 & .717 ± .024 & N/A & N/A & .579 ± .007 & .618 ± .009\\
& BiLSTM* - EMD+SF & N/A & N/A & .718 ± .007 & .773 ± .006 & .557 ± .016 & .593 ± .017\\

\cdashline{2-8}[2pt/3pt]
& MGADE - TRC +EMD & .739 ± .022 & .710 ± .015 & .689 ± .007 & .775 ± .005 & N/A & N/A\\
& MGADE - TRC +SF & .750 ± .017 & .714 ± .013 & N/A & N/A & .528 ± .037 & .603 ± .030\\

\midrule
\multirow{10}{*}{BSC}

& RoBERTa* - EMD & N/A & N/A & .691 ± .004 & .773 ± .008 & N/A & N/A\\
& RoBERTa* - SF & N/A & N/A & N/A & N/A & .541 ± .007 & .615 ± .009\\
& RoBERTa* - TRC\textsuperscript{$\#$}+EMD & .779 ± .033 & .741 ± .023 & \textbf{.716 ± .012} & \textbf{.787 ± .012} & N/A & N/A\\
& RoBERTa* - TRC\textsuperscript{$\#$}+SF & \textbf{.804 ± .008} & \textbf{.757 ± .005} & N/A & N/A & \textbf{.585 ± .002} & \textbf{.657 ± .006}\\
& RoBERTa* - EMD+SF & N/A & N/A & .708 ± .002 & .781 ± .007 & .561 ± .007 & .627 ± .009\\

\cdashline{2-8}[2pt/3pt]
& BiLSTM* - EMD & N/A & N/A & .677 ± .008 & .753 ± .011 & N/A & N/A\\
& BiLSTM* - SF & N/A & N/A & N/A & N/A & .520 ± .011 & .583 ± .009\\
& BiLSTM* - TRC\textsuperscript{$\#$}+EMD & .751 ± .047 & .714 ± .030 & .704 ± .006 & .773 ± .007 & N/A & N/A\\
& BiLSTM* - TRC\textsuperscript{$\#$}+SF & .755 ± .007 & .717 ± .008 & N/A & N/A & .566 ± .007 & .624 ± .008\\
& BiLSTM* - EMD+SF & N/A & N/A & .698 ± .005 & .770 ± .007 & .544 ± .016 & .599 ± .016\\
\cdashline{2-8}[2pt/2pt]
& MGADE - TRC\textsuperscript{$\#$} +EMD & .707 ± .025 & .687 ± .015 & .658 ± .014 & .759 ± .020 & N/A & N/A\\
& MGADE - TRC\textsuperscript{$\#$} +SF & .732 ± .030 & .705 ± .019 & N/A & N/A & .527 ± .016 & .598 ± .017\\

\midrule
\multirow{12}{*}{Ground Truth}

& RoBERTa - TRC & .833 ± .006 & .773 ± .011 & N/A & N/A & N/A & N/A\\
& RoBERTa* - EMD & N/A & N/A & .735 ± .008 & .803 ± .007 & N/A & N/A\\
& RoBERTa* - SF & N/A & N/A & N/A & N/A & .617 ± .002 & .697 ± .011\\
& RoBERTa* - TRC+EMD & \textbf{.847 ± .008} & \textbf{.795 ± .008} & \textbf{.752 ± .006} & \textbf{.815 ± .009} & N/A & N/A\\
& RoBERTa* - TRC+SF & .838 ± .013 & .779 ± .014 & N/A & N/A & \textbf{.648 ± .003} & \textbf{.705 ± .014}\\
& RoBERTa* - EMD+SF & N/A & N/A & .749 ± .012 & .811 ± .013 & .646 ± .003 & .705 ± .008\\

\cdashline{2-8}[2pt/3pt]
& BiLSTM - TRC & .774 ± .015 & .674 ± .012 & N/A & N/A & N/A & N/A\\
& BiLSTM* - EMD & N/A & N/A & .727 ± .004 & .787 ± .004 & N/A & N/A\\
& BiLSTM* - SF & N/A & N/A & N/A & N/A & .582 ± .007 & .621 ± .007\\
& BiLSTM* - TRC+EMD & .817 ± .009 & .743 ± .013 & .747 ± .006 & .800 ± .003 & N/A & N/A\\
& BiLSTM* - TRC+SF & .814 ± .008 & .748 ± .012 & N/A & N/A & .620 ± .013 & .655 ± .010\\
& BiLSTM* - EMD+SF & N/A & N/A & .746 ± .005 & .797 ± .008 & .602 ± .008 & .633 ± .013\\
\cdashline{2-8}[2pt/3pt]
& MGADE - TRC +EMD & .826 ± .019 & .770 ± .016 & .714 ± .012 & .789 ± .012 & N/A & N/A\\
& MGADE - TRC +SF & .810 ± .029 & .732 ± .044 & N/A & N/A & .562 ± .016 & .636 ± .029	\\

		\bottomrule
		\end{tabular}
		\end{adjustbox}
    \vspace{8pt}

}  
\end{adjustbox}
  \caption{Performance of each method learning from noisy labels and ground truth labels on three tasks of text relevance classification (TRC), entity mention detection (EMD), and slot filling (SF). Metrics: Accuracy and ${F_1}$ for TRC, Strict ${F_1}$ and Entity Type ${F_1}$ for EMD and SF. *: with a CRF layer. \textsuperscript{$\#$}: the labels are from majority voting (MV).}
  \label{tab:main_results}
  \vspace{-10pt}
\end{table*}

Based on the labels collected, we design three tasks and 
evaluate state-of-the-art deep learning methods on these three tasks with both the \textbf{aggregated label} and the \textbf{ground truth label}. 

{\bf Text relevance classification (TRC).}
This task is to predict whether a given tweet indicates a possible foodborne illness incident. As described in Section \ref{sec:aggregated}, in our dataset, each tweet comes with a binary class label, which denotes whether the tweet is relevant to foodborne illness or not. These binary class labels are the gold standard labels for TRC tasks. Models trained on this task can be applied to detect possible foodborne illness incidents mentioned in online posts.

{\bf Entity mention detection (EMD).}
In our data set, we highlight all words or phrases belonging to specific entity classes (food, location, symptom, and foodborne illness keywords). The EMD task aims to identify \textbf{all} the mentions of the four types of entities listed above. EMD is different from the classical Named Entity Recognition (NER) task, in that the latter only focuses on the named entities. Here, EMD requires extracting both named and unnamed entities. Actually, in many tweets, people may mention many entities without providing their actual names. Therefore, the EMD task may help us find more information related to foodborne illness incidents than the NER task.

{\bf Slot filling (SF).}
This task is to extract the text spans that can fill the slot defined in Table \ref{tab:slot_def} in a given tweet. As described in the last paragraph, entity mentions that play specific roles in a foodborne illness incident contain more valuable information than other entity mentions. 

{\bf Combination Tasks.}
In our setting, a tweet contains one or more slots if and only if the tweet is labeled as a relevant tweet. These slots are always a subset of all entities. In this study, in addition to learning each task individually, we also design several combinations of these tasks that can be learned jointly to observe multi-task models' performance on these tasks. The first two are TRC + EMD and TRC + SF. We pair TRC with the other two tasks since we want to see if the sentence-level and token-level can benefit from each other. Especially the SF task, given that in our dataset, at least one entity mention must be a slot value if and only if the tweet indicates a possible foodborne illness incident. The model that can utilize such relationships may achieve better performance. We also introduce the combination of EMD and SF tasks to observe the relation between the two token tasks.

\subsection{Models}
To establish a starting point for future study, we evaluate several baseline models on TWEET-FID. Configurations of each method and implementation details are available in the Appendix.

\begin{itemize}
\item {RoBERTa:  This pretrained deep learning model, proposed by \cite{liu2019roberta},  builds on BERT \cite{devlin2018bert}, and refines the pre-training procedure by removing the next-sentence prediction tasks and training with larger learning rates and mini-batch. \cite{liu2019roberta} implements RoBERTa on multiple tasks and shows that RoBERTa can achieve state-of-the-art results on several benchmarks. RoBERTa is implemented for all tasks. For each task, we pair RoBERTa with a classification head as the top layer (linear layer on top of hidden state output). For EMD and SF tasks, we create the network with a CRF layer on top of the linear classification layer. For multi-task combinations (TRC + EMD, TRC + SF), we assign two classification heads on top of the model's hidden state output, one head for one task. }


\item {BiLSTM:
Bidirectional LSTM is a widely used sequence architecture. In this study, we input the sequence of GloVe embeddings\footnote{We used the pre-trained word embedding of Common Crawl which includes 840B tokens and 2.2M vocabularies. \url{https://nlp.stanford.edu/projects/glove/}} of each tweet into the BiLSTM, the hidden states of the BiLSTM are fed into a classifier for the final prediction. Same as RoBERTa, we add a CRF layer on top of the linear classification layer for EMD and SF tasks. For multi-task combinations (TRC + EMD, TRC + SF), we assign two classification heads on top of the BiLSTM's hidden states, one head for one task. The TRC task classification layer takes the concatenation of the last hidden states from both directions as input. EMD and SF tasks take the hidden state of each word from both directions as input.}


\item {MGADE: A multi-grained joint deep network model proposed by \cite{wunnava-etal-2020-dual},  was designed to concurrently solve both ADE entity recognition and ADE assertive sentence classification. The model is equipped with a dual-attention mechanism to construct multiple distinct sentence representations to capture both task-specific and semantic information in a sentence. As for the BiLSTM model, the pre-trained GloVe embedding is provided to the model. We applied this model to our TRC + EMD and TRC + SF tasks.}



\end{itemize}

\subsection{Experimental results}

Table \ref{tab:main_results} presents the evaluation results of each method. When comparing single-task and multi-task methods, we observe that multi-task methods can match or exceed the performance of single-task counterparts. EMD and SF achieve the highest gain in $F_1$ scores when they are jointly learned with the TRC task. Additionally, the best performance of the TRC task is reached when it is jointly learned with token-level tasks. It illustrates that the sentence-level and the token-level task can mutually benefit each other. When EMD and SF are jointly learned, performance on two tasks also improves compared to single-task methods. These findings suggest that two token-level tasks can also benefit each other.

When comparing the performance of the methods on the EMD and SF tasks, we notice that methods reach higher $F_1$ scores on the EMD task. The main reason is that in the SF task, a word, for example, \textit{sandwich}, can only fill a slot when related to a foodborne illness incident, otherwise labeled as ''out of slot''. But in the EMD task, in most cases, \textit{sandwich} is labeled as a food entity regardless of the relevance between the word and a foodborne illness incident.
It makes the SF task harder than the EMD task since the relevant information has to be incorporated when making predictions on the SF task.  We also observe that RoBERTa-based methods outperform other methods on many tasks. This is likely because RoBERTa is pre-trained on a large corpus, whereas MGADE and BiLSTM are being trained from scratch on our dataset. Another finding is that methods learned from ground truth labels perform much better than those learned from aggregated noisy labels. The performance gaps on the SF task are much larger than those on the EMD and TRC task. This finding is consistent with the evaluation results of the aggregated labels for the training and validation set in Table \ref{tab:aggregated_eval}. Since aggregated slot type labels are least close to ground truth labels, methods trained on such poorly labeled data should also perform worse when evaluating on the ground truth test labels.


\section{Conclusion}
In this work, we construct TWEET-FID, the first publicly-available tweet dataset for multiple foodborne illness incident detection tasks. TWEET-FID serves as training data for the following tasks: (1) identify if a given tweet indicates a foodborne illness incident; (2) find and extract entities in the tweet; (3) only extract entities related to the foodborne illness incident. We conduct a crowdsourcing study to collect annotations for these three tasks. 
Thereafter, we conduct an experimental study to evaluate the annotation aggregation methods and generate the consensus label for the three tasks mentioned above. Moreover, we evaluate state-of-the-art single-task and multi-task models for realizing the three tasks and investigate the performance of learning from weak labels. These results can serve as the baseline for future research work on label aggregation, weakly supervised learning, and multitask learning not only in this important area of food safety but also in general NLP domain.

\section{Acknowledgement}
This work was supported by the Agriculture and Food Research Initiative (AFRI) award no. 2020-67021-32459 from the U. S. Department of Agriculture (USDA) National Institute of Food and Agriculture (NIFA) and the Illinois Agricultural Experiment Station. We thank Mengrui Luo for her assistance in implementing the crowdsourcing interface for this study.

\section{References}
\label{lr:ref}

\bibliographystyle{lrec2022-bib}
\bibliography{ref}

\newpage
\appendix

\section{Appendix}
\subsection{Statistics of entities and slots for the aggregated labels}
\begin{table}[!htb]

\begin{adjustbox}{width=\linewidth,center}
{

    \centering
    
    \begin{adjustbox}{width=\linewidth}
        \begin{tabular}{l  c c c c r}
        \toprule
        \  &  & Train & Validation & Test & Full Set \\
		\midrule[\heavyrulewidth]
\multirow{5}{*}{Entity}
&Food   & 1032 & 133 & 125 & 1209  \\
&Location   & 681     & 72         & 80      & 833       \\
&Symptom   & 1160     & 147         & 157      & 1464   \\
&Other  & 2324     & 292         & 288      & 2904 \\
\cmidrule{2-6}
& \textbf{Total}  & \textbf{5197} & \textbf{644} & \textbf{650} & \textbf{6491}    \\
\midrule
\multirow{5}{*}{Slot}
& What & 408 & 47 &61 & 516 \\
& Where & 432 & 44 & 53 & 529 \\
& Symptom & 377 & 43 & 60 & 480 \\
& Key & 1117 & 142 & 140 & 1399 \\
\cmidrule{2-6}
& \textbf{Total}  & \textbf{2334} & \textbf{276} & \textbf{314} & \textbf{2924}    \\

		\bottomrule
		\end{tabular}
		\end{adjustbox}

}  
\end{adjustbox}

  \caption{Statistics of entity types and slot types for the  expert labels\\ }
  \label{tab:expert_label_stat}
  \vspace{-10pt}
\end{table}
\begin{table}[!htb]

\begin{adjustbox}{width=\linewidth,center}
{

    \centering
    
    \begin{adjustbox}{width=\linewidth}
        \begin{tabular}{p{1.5cm}l  c c c}
        \toprule
        \  &  & Train & Validation  \\
		\midrule[\heavyrulewidth]
\multirow{5}{*}{Entity}
&Food   & 948 & 121   \\
&Location   & 746     & 78                \\
&Symptom   & 1103    & 138            \\
&Other  & 2294     & 292         \\
\cmidrule{2-4}
& \textbf{Total}  & \textbf{5091} & \textbf{629}     \\
\midrule
\multirow{5}{*}{Slot}
& What & 535 & 56  \\
& Where & 447 & 47  \\
& Symptom & 608 & 71  \\
& Key & 1599 & 211  \\
\cmidrule{2-4}
& \textbf{Total}  & \textbf{3189} & \textbf{385}     \\

		\bottomrule
		\end{tabular}
		\end{adjustbox}

}  
\end{adjustbox}

  \caption{Statistics of entity types and slot types for the majority voting labels\\ }
  \label{tab:mv_label_stat}
  \vspace{-10pt}
\end{table}
\begin{table}[!htb]

\begin{adjustbox}{width=\linewidth,center}
{

    \centering
    
    \begin{adjustbox}{width=\linewidth}
        \begin{tabular}{l  c c c}
        \toprule
        \  &  & Train & Validation  \\
		\midrule[\heavyrulewidth]
\multirow{5}{*}{Entity}
&Food   & 1072 & 139   \\
&Location   & 898     & 96               \\
&Symptom   & 1439     & 175            \\
&Other  & 2412     & 298         \\
\cmidrule{2-4}
& \textbf{Total}  & \textbf{5821} & \textbf{708}     \\
\midrule
\multirow{5}{*}{Slot}
& What & 586 & 62  \\
& Where & 549 & 67 \\
& Symptom & 750 & 88  \\
& Key & 1649 & 209  \\
\cmidrule{2-4}
& \textbf{Total}  & \textbf{3534} & \textbf{426}     \\

		\bottomrule
		\end{tabular}
		\end{adjustbox}

}  
\end{adjustbox}

  \caption{Statistics of entity types and slot types for the   BSC aggregated labels\\ }
  \label{tab:bsc_label_stat}
  \vspace{-10pt}
\end{table}

\subsection{Experiment implementation details}
We implement the code of BSC-seq \footnote{\url{https://github.com/UKPLab/arxiv2018-bayesian-ensembles}} to aggregate crowdsourced entity sequence data. The code is licensed under Apache-2.0 License. 

We use the pre-trained RoBERTa-base model from the ``Transformers" library\footnote{\url{https://github.com/huggingface/transformers}}, which is licensed under Apache-2.0 License. For MGADE, we used the implementation from \url{https://github.com/swunnava20/mgade}, which is licensed under MIT License. All experiments are run on a A100 GPU. 

 For every task, we report the average evaluation results on the test dataset of each model from 5 random initialization. RoBERTa-based and BiLSTM-based methods are trained with 20 epochs. We used the Adam optimizer for RoBERTa and BiLSTM. We searched for best learning rate for RoBERTa-based and BiLSTM-based method in the set of \{$1\time10^{-5}$, $1\time10^{-4}$, $1\time10^{-3}$\} based on results on validation set. After searching, we determine the initial learning rate of RoBERTa is $1\time10^{-5}$, for BiLSTM, the learning rate is $1\time10^{-3}$.

When implementing RoBERTa/BiLSTM methods on TRC + EMD/SF tasks, we define a joint loss function in the training process upon the losses specified for different sub-tasks as follows: 
\begin{equation}
\label{eqn:joint_loss}
    J(\theta) = J_{TRC} +  \lambda J_{EMD/SF}.
\end{equation}
Here, $J_{TRC}$ and $J_{EMD/SF}$ are cross-entropy losses. We introduce a hyper-parameter $\lambda$ into the joint loss to strike a balance between TRC and EMD/SF task. We search for best $\lambda$ from the set of \{0.1, 1, 10, 100\} on validation set. After searching, we found the best $\lambda$ is 10. For the MGADE method, the learning rate is $1\times10^{-3}$. For EMD + SF task, we define a similar joint loss function without $\lambda$ term.

More details about model configurations are available at our code repository\footnote{\url{https://github.com/ruofanhu/Tweet-FID}}.

\subsection{License}
The person in request (“the user”) may receive and use TWEET-FID (“the dataset”) only after accepting and agreeing to both the Twitter Terms of Service, Privacy Policy, Developer Agreement, and Developer Policy and the following terms and conditions: 

\textbf{Commercial and academic use}

The dataset is made available for non-commercial purposes only. Any commercial use of this data is forbidden. 

\textbf{Redistribution }

The user is not allowed to copy and distribute the dataset or parts of it to a third party without first obtaining permission from the creators. 

\textbf{Publications}

The use of data for illustrative purposes in publications is allowed. Publications include both scientific papers and presentations for scientific/educational purposes. 

\textbf{Citation}

All publications reporting on research using this dataset have to acknowledge this by citing the following article: 

Ruofan Hu, Dongyu Zhang, Dandan Tao, Thomas Hartvigsen, Hao Feng, Elke Rundensteiner, \textit{“TWEET-FID: An Annotated Dataset for Multiple Foodborne Illness Detection Tasks”}, in Submission at the 13th Conference on Language Resources and Evaluation (LREC 2022).

For specific software output that is shared as part of this data, the user agrees to respect the individual software licenses and use the appropriate citations as mentioned in the documentation of the data. 

\textbf{TWEET-FID changes}

The creators of this dataset are allowed to change these terms of use at any time. In this case, users will have to accept and agree to be bound by new terms to keep using the dataset. 

\textbf{Warranty}

The dataset comes without any warranty. In no event shall the provider be held responsible for any loss or damage caused using this data.

\end{document}